\documentclass[letterpaper, 10 pt, conference]{ieeeconf}
\IEEEoverridecommandlockouts    %
\usepackage{graphics}           
\usepackage{times}              
\usepackage{amsmath}            
\usepackage{amssymb}            
\usepackage{graphicx}
\usepackage{algorithm}
\usepackage[noend]{algpseudocode}
\usepackage{booktabs}
\usepackage{color}
\usepackage{siunitx}
\definecolor{instructioncolor}{rgb}{.5,.5,.5}

\usepackage[font=small]{caption}

\def\secref#1{Sec.~\ref{#1}}
\def\figref#1{Fig.~\ref{#1}}
\def\tabref#1{Tab.~\ref{#1}}
\def\eqref#1{Eq.~(\ref{#1})}

\makeatletter
\usepackage{xspace}
\DeclareRobustCommand\onedot{\futurelet\@let@token\@onedot}
\def\@onedot{\ifx\@let@token.\else.\null\fi\xspace}

\def\etal{{et al}\onedot}
\makeatother

\usepackage{array}
\newcolumntype{L}[1]{>{\raggedright\let\newline\\\arraybackslash\hspace{0pt}}m{#1}}
\newcolumntype{C}[1]{>{\centering\let\newline\\\arraybackslash\hspace{0pt}}m{#1}}
\newcolumntype{R}[1]{>{\raggedleft\let\newline\\\arraybackslash\hspace{0pt}}m{#1}}

\def\argmax{\mathop{\rm argmax}}

\newcommand{\norm}[1]{\lVert#1\lVert}

\newcommand{\degrees}{{\mbox{$^\circ$}}}

\usepackage{tabularx, graphicx, amsmath, amsfonts, pifont, amssymb, subcaption, balance}

\usepackage{todonotes}

\title{\LARGE \bf Markerless Aerial-Terrestrial Co-Registration of Forest Point Clouds using a Deformable Pose Graph}

\author{Beno\^{i}t Casseau$^{1}$ \and Nived Chebrolu$^{1}$ \and Matias
Mattamala$^{1}$ \and Leonard Freissmuth$^{1,2}$ \and Maurice
Fallon$^{1}$%
\thanks{$^{1}$The authors are with the University of Oxford, UK.
\texttt{\{benoit, nived, matias, mfallon\}@robots.ox.ac.uk }.} \thanks{$^{2}$The
author is also with the Technical University of Munich, Germany.
\texttt{l.freissmuth@tum.de} } }

\begin{document}
\maketitle
\thispagestyle{empty}
\pagestyle{empty}

\begin{abstract}
  For biodiversity and forestry applications, end-users desire maps of forests
  that are fully detailed---from the forest floor to the canopy. Terrestrial
  laser scanning and aerial laser scanning are accurate and increasingly mature
  methods for scanning the forest. However, individually they are not able to
  estimate attributes such as tree height, trunk diameter and canopy density due
  to the inherent differences in their field-of-view and mapping processes.   In
  this work, we present a pipeline that can automatically generate a single
  joint terrestrial and aerial forest reconstruction. The novelty of the
  approach is a marker-free registration pipeline, which estimates a set of
  relative transformation constraints between the aerial cloud and terrestrial
  sub-clouds without requiring any co-registration reflective markers to be
  physically placed in the scene. Our method then uses these constraints in a
  pose graph formulation, which enables us to finely align the respective clouds
  while respecting spatial constraints introduced by the terrestrial SLAM
  scanning process.  We demonstrate that our approach can produce a fine-grained
  and complete reconstruction of large-scale natural environments, enabling
  multi-platform data capture for forestry applications without requiring
  external infrastructure.
\end{abstract}

\section{Introduction}
\label{sec:intro}

Terrestrial laser scanning (TLS, tripod-based) and mobile laser scanning (MLS,
human or robot carried) play an emerging role in biodiversity monitoring and
forestry by capturing the detailed geometric data required for the study of
individual trees.  
These data can be used to recover highly detailed reconstructions of the
under-canopy, which can be used to extract traits such as the
diameter-at-breast-height (DBH), tree species and general forest
density~\cite{white2016remote},\cite{murtiyoso2024}. Such traits can be used for
decision-making in forestry as well as robotic forest
harvesting~\cite{jelavic2021autonomous}. However, TLS/MLS data alone are not
able to capture other traits such as the tree height or crown volume, which
cannot be accurately observable from the ground due to limitations in the sensor
range as well as blocking foliage and other occlusions~\cite{wang2019situ}.

Aerial vehicles, such as MAVs and fixed-wing aircraft, fly above the canopy to
carry out aerial laser scanning (ALS). These data captures structural details
that cannot be obtained from terrestrial scanning. ALS survey can also
reconstruct much larger areas in a single flight (tens of hectares).
Nevertheless, they intrinsically produce lower density reconstructions of the
understory than TLS or MLS, making it difficult to determine stem properties in
dense forests~\cite{wang2019situ}. Given the complementary nature of the sensing
modalities, combining the terrestrial and aerial reconstructions would allow
comprehensive characterization of tree and forest structures in dense and
diverse ecosystems.
A further advantage of the combined map is that it serves as an effective reference map, improving robot localization capabilities.

\begin{figure}[t]
  \centering
  \includegraphics[width=\linewidth]{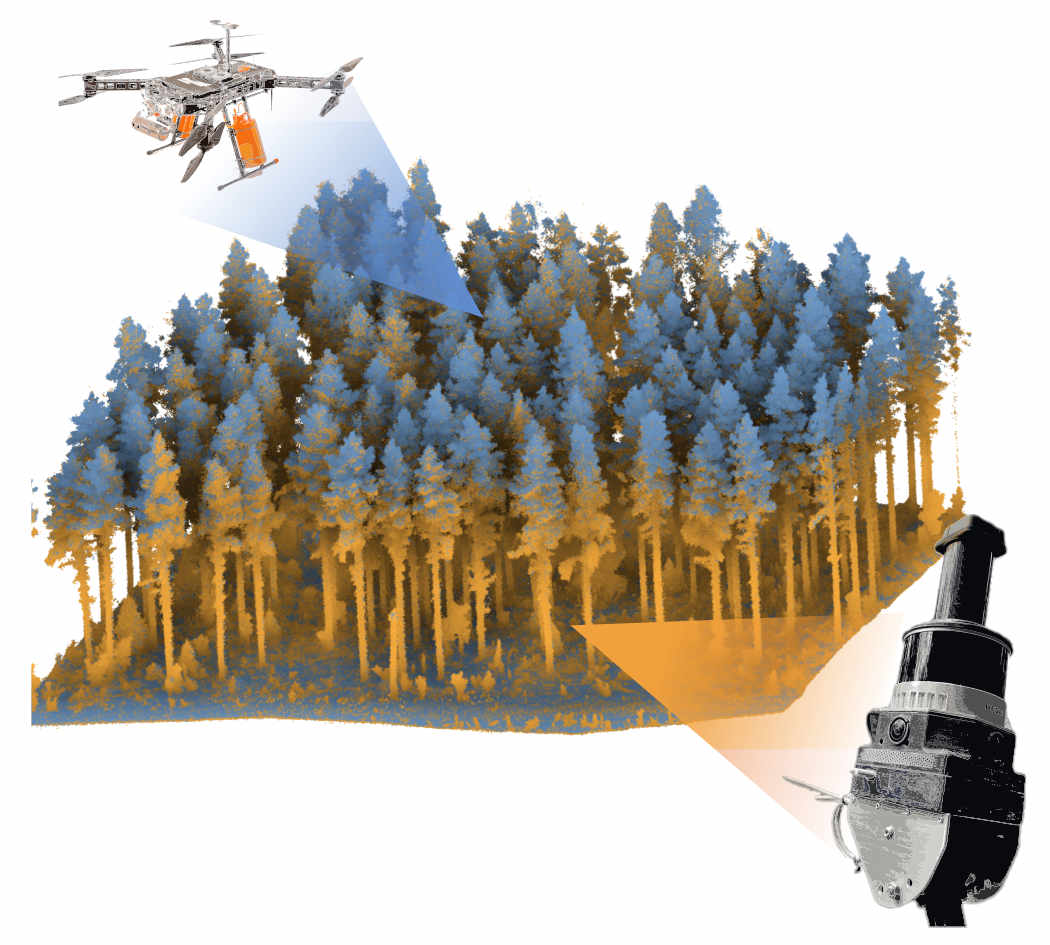}
  \caption{A forest map built using a terrestrial map from our mobile scanning
  system (orange) and an aerial map from a drone (blue). Together, the scans
  produce a map reconstructing the canopy and understory in sufficient detail
  for individual-tree forest inventory.}
  \vspace{-15pt}
  \label{fig:motivation}
\end{figure}

\begin{figure}[htbp]
  \centering
  \includegraphics[width=\linewidth]{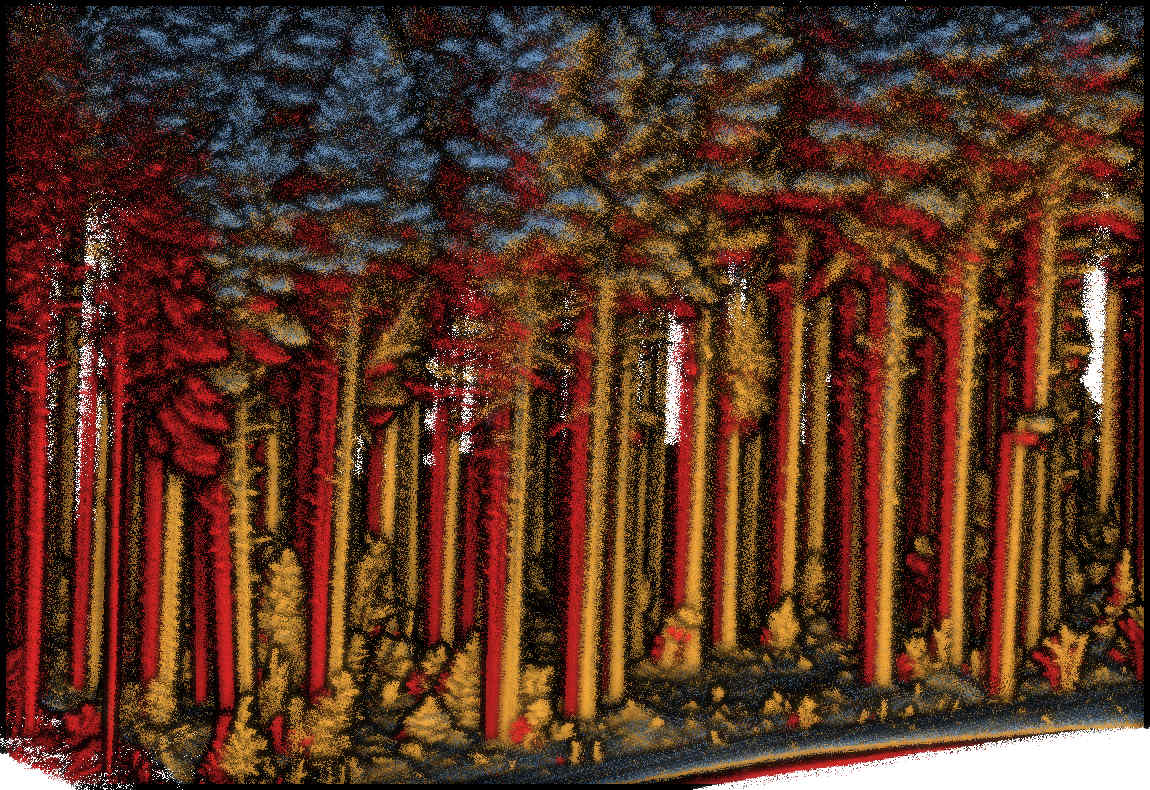} 
  \caption{An unoptimized MLS point cloud (in red) is poorly aligned with the ALS map (blue). On the other hand, the optimized cloud (in orange) is directly aligned with the ALS map. The shearing in the red MLS point cloud is due to the small drift in the SLAM system (a few meters across a kilometer scale map).}
  \label{fig:mls_recon_errors}
\end{figure}

The standard way of achieving joint ALS-MLS reconstruction is to manually place
reflective markers in the forest which can be detected by both ground and aerial
systems. This facilitates the alignment of the clouds, either through manual
intervention or automated processes~\cite{hilker2012simple}. This method
involves a time-consuming setup of markers, and ensuring marker visibility for
both ALS and MLS systems is challenging in dense forests. Hence, there is an
interest in developing marker-free registration methods, with recent
advancements showing promise~\cite{shao2022arxiv,castorena2023arxiv}. These
methods perform a rigid alignment of the aerial and terrestrial clouds, assuming
that both are independently accurate. However, while uniform accuracy can be
guaranteed for ALS thanks to the fusion of inertial measurement units (IMU) and
global navigation satellite system (GNSS); a mobile scanner cannot rely on GNSS
under canopy due to occlusion and signal loss. Therefore, MLS typically
estimates their pose via a SLAM system using onboard LiDAR and IMU measurements
with a crude \SIrange{5}{10}{\meter} accurate GNSS position estimate available.
While MLS (from onboard LiDAR odometry) can produce accurate local
reconstruction, it can still suffer from drift in long sequences
(\figref{fig:mls_recon_errors}).

In this paper, we present an approach to combine aerial and terrestrial data to
achieve high-fidelity georeferenced point cloud reconstructions by automatically
finding matches between individual, local terrestrial sub-maps and a single
global, georeferenced ALS map without requiring specialized markers. The
approach combines global aerial maps with the locally accurate MLS sub-maps, by
jointly optimizing them in a factor graph-based formulation that combines local
and global constraints. We demonstrate that our method enables denser, complete
reconstructions of the full trees, which enable further forestry analyses, as
shown in \figref{fig:motivation}.

The specific contributions of our work are:
\begin{itemize}
\item A method to compute local aerial-terrestrial registration that does not
require custom targets for alignment, only requiring a rough GNSS prior position
estimate.
\item A pose graph-based formulation that performs fine-grained alignment of
terrestrial clouds to a global aerial reference, applicable to both sequences
captured with MLS as well as a set of individual TLS scans.
\item Experimental validation using aerial and terrestrial datasets obtained in
a Swiss mixed forest and a Finnish conifer forest.
\end{itemize}

\section{Related Work}
\label{sec:related}
The problem of aerial and terrestrial cloud registration has been motivated by
the different scanning perspectives provided by each
technique~\cite{chasmer2004assessing}. Early approaches achieved this by manual
alignment of the different different sources, which was generally assisted by
some automatic components. Lovell et al.~\cite{lovell2003aerial} presented one
of the earliest examples of measuring canopy in an Australian forest by
combining both data sources. Omasa et al.~\cite{omasa2008three} also combined
aerial and terrestrial scans to model an urban park and trees using rendered
bird-eye views. More recent approaches have used automatic tree detection
instead but still required manual registration~\cite{kankare2014accuracy}.

To achieve semi-automatic matching, specialized markers (e.g. reflective
targets) have been used for the co-registration of independent
clouds~\cite{hilker2012simple,ge2021target}. Chasmer et
al.~\cite{chasmer2004assessing} used targets to study the distribution of scans
captured by ground and aerial sensing. Similarly, Holopainen et
al.~\cite{holopainen2013tree} used reflective spheres to align clouds obtained
in urban forests. Nevertheless, these solutions require careful positioning of
the markers to maximize the co-visibility in aerial and terrestrial scans.

Instead, we are interested in automatic solutions, which do not require any
specialized markers to perform the co-registration. This is generally achieved
by having an aerial-terrestrial matching step, and then a registration procedure
that provides a relative transformation between aerial and terrestrial scans.
For matching, previous approaches have explored strategies such as GNSS
priors~\cite{haughlin2014geo}, as well as identifying matching features between
the scans, including tree density analysis~\cite{dai2019automated}, tree crown
analysis~\cite{paris2017automatic} or stem position
matching~\cite{bienert2009methods}. In this work we leverage some of the
previous techniques, particularly detecting tree stems, to deal with partial
scans due to the incremental nature of MLS scanning.

Different registration approaches have been explored depending on the
application domain. For forestry applications where the objective is to provide
denser reconstructions, previous works have explored techniques such as graph
matching to provide a registration solution~\cite{polewski2019markerfree}, which
can be further refined with standard point cloud alignment algorithms such as
iterative closest point (ICP)~\cite{BeslM92}. Alternatively, non-linear
optimization refinement has been proposed to achieve fine-grained
alignment~\cite{dai2019automated, castorena2023arxiv}. Other solutions based on
particle filtering have been proposed for related problems such as robot
localization using an aerial map~\cite{delima23ral}, which are beyond the scope
of this work.

Our system builds upon the previous ideas to combine some of the co-registration
techniques with an optimization-based refinement step, similarly as done by
Castorena et al.~\cite{castorena2023arxiv}. However, we explicitly aim to
introduce spatial consistency in the refinement using a factor graph
formulation, which is \emph{deformed} to better align the aerial scan.

\section{Method}
\label{sec:main}
\begin{figure}[t]
  \centering
   \includegraphics[width=0.95\linewidth]{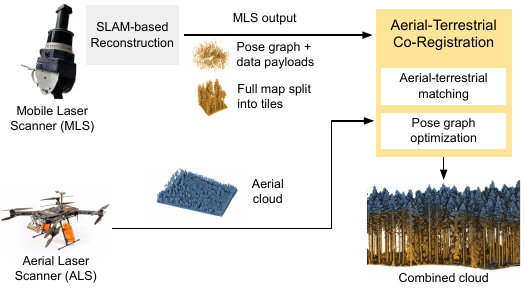}
  \caption{General system overview. Our aerial-terrestrial co-registration system combines the aerial clouds from ALS, with a terrestrial reconstruction obtained by means of a LiDAR-Inertial odometry and pose-graph SLAM system.}
  \label{fig:system-overview}
\end{figure} 

In this section, we present our aerial-terrestrial co-registration pipeline,
shown in \figref{fig:system-overview}. The inputs are an aerial point cloud in
UTM frame (\emph{ALS cloud}), and terrestrial scans from MLS (\emph{MLS
clouds}). The MLS clouds can be in either a \emph{pose graph format} or a
\emph{tile format}. 

\begin{figure}
  \centering
  \begin{subfigure}[b]{0.42\columnwidth}
      \centering
      \includegraphics[width=4cm, height=3.6cm]{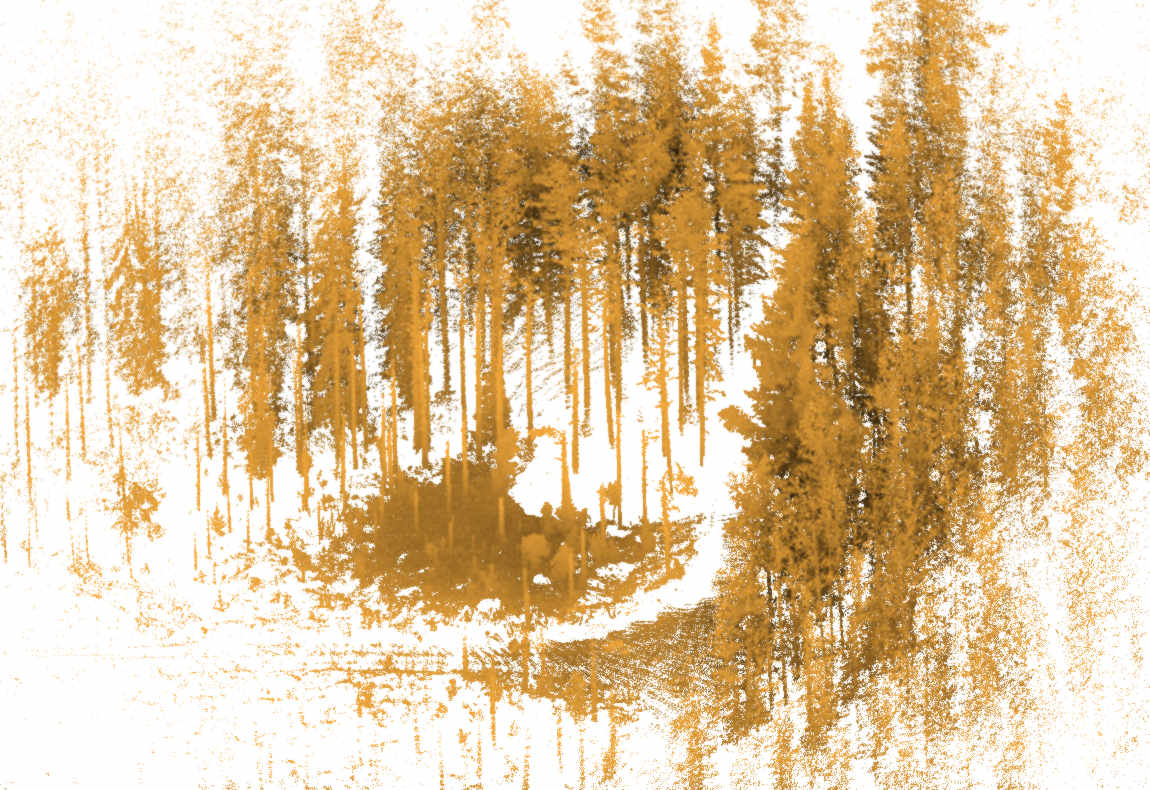}
      \caption{Data payload}
  \end{subfigure}
  \hspace{0.2cm}
  \begin{subfigure}[b]{0.42\columnwidth}
      \centering
      \includegraphics[width=\textwidth, height=3.6cm]{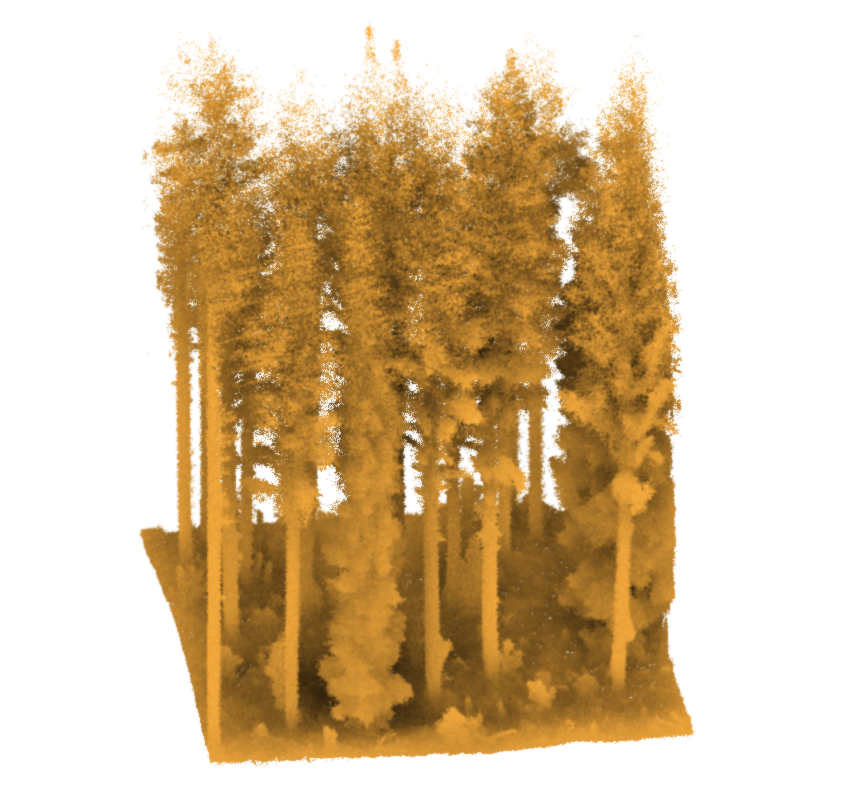}
      \caption{Tile cloud}
  \end{subfigure}
  \caption{We consider two formats for the MLS data. The \emph{pose graph
  format} is defined by a pose graph and a collection of data payloads (a),
  obtained by temporal aggregation of consecutive LiDAR scans; payloads cover a
  larger area but are sparse and lack canopy points. The \emph{tile format}
  represents the MLS mission as a collection of tiles (b), obtained by
  aggregating all the mission scans into a single, dense cloud and partitioning
  them into a fixed-size grid.}
  \label{fig:mls-format}
  \vspace{-1mm}
\end{figure}

In \textit{pose graph format} the MLS clouds are represented by a SLAM pose
graph in UTM frame with local terrestrial point clouds attached to each pose.
The clouds---which we name \emph{data payloads}---are obtained by temporally
aggregating LiDAR scans at sensor rate (\SI{10}{\hertz}) within a fixed
distance; an example is shown in \figref{fig:mls-format} (a). 
The data payload is converted from a local coordinate frame to the UTM frame
using the GNSS measurements acquired by the on-board receiver. This pose graph
format enables fine-grained aerial-terrestrial alignment by exploiting the
temporal information about how the data was acquired. 

In \textit{tile format} we consider the full, single cloud obtained by joining
all the data payloads acquired over a mission and then partitioning the data
into a fixed-size, gravity-aligned 2D grid to produce a collection of
\emph{tiles}. We use a tile of size \SI{20}{\meter} $\times$ \SI{20}{\meter} in
this work, as shown in \figref{fig:mls-format} (b). This representation is
independent of the MLS device, and does not require access to the SLAM
information---only the relative transformation between the tiles specified by
the grid.

\subsection{Aerial-terrestrial Matching}
\label{subsec:matching}

\begin{figure}[t]
  \centering
  \includegraphics[width=0.95\linewidth]{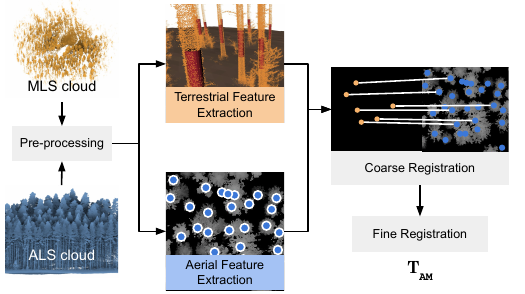}
  \caption{The aerial-terrestrial registration step aims to find the relative
  transformation between a large aerial and a much smaller local terrestrial
  cloud, $\mathbf{T}_{\texttt{AM}}$. }
  \label{fig:registration-pipeline}
\end{figure} 

The first step aims to find a relative transformation between each MLS cloud and
the aerial cloud (\figref{fig:registration-pipeline}). The procedure is
independent of the MLS cloud format.

We denote $\mathbf{T}_{\texttt{AM}}$ the transformation that maps an individual
MLS cloud (payload/tile) in $\texttt{M}$ frame into the ALS frame $\texttt{A}$.
The set of relative transformations obtained for each of the MLS clouds will
provide global constraints for the pose graph optimization procedures presented
later in \secref{subsec:pose-graph}.

\subsubsection{Pre-processing}
Since the ALS cloud can be very large
($\sim$\SI{600}{\meter}$\times$\SI{600}{\meter} in our datasets), we reduce the
search space by cropping the ALS cloud around a neighborhood of the center of
the MLS cloud. This is achieved by using the approximate GNSS position of the
MLS cloud (accurate to about \SIrange{5}{10}{\meter}) and a padding factor
depending on the MLS cloud's size. On average, this produced crops of
\SI{30}{\meter}$\times$\SI{30}{\meter} for both payloads and tiles in our
experiments.

While the MLS clouds are gravity-aligned and approximately aligned with North
(as with the ALS cloud), we observe a height offset compared to the ALS cloud.
This may be attributed to an imprecise GNSS fix obtained by the MLS under the
forest canopy. We correct this misalignment of the clouds along the height axis
by matching the ground planes from both the MLS and ALS clouds. To rectify this
misalignment along the height axis, we estimate the correction by matching the
ground planes derived from both the MLS and ALS clouds. This process involves
identifying ground points based on their normal information and employing a
RANSAC procedure to fit a plane to each MLS cloud as well as the ALS cloud.
Using the fitted planes from ALS and MLS clouds,  
we determine a rotation correction by aligning the normal vectors to the plane
as well as the vertical offset between the ALS and terrestrial MLS clouds,
similar to the method by Shao~\etal~\cite{shao2022arxiv}.

\subsubsection{Aerial Feature Extraction}
\begin{figure}
  \centering
  \begin{subfigure}[b]{0.42\columnwidth}
      \centering
      \includegraphics[width=\textwidth, height=3.6cm]{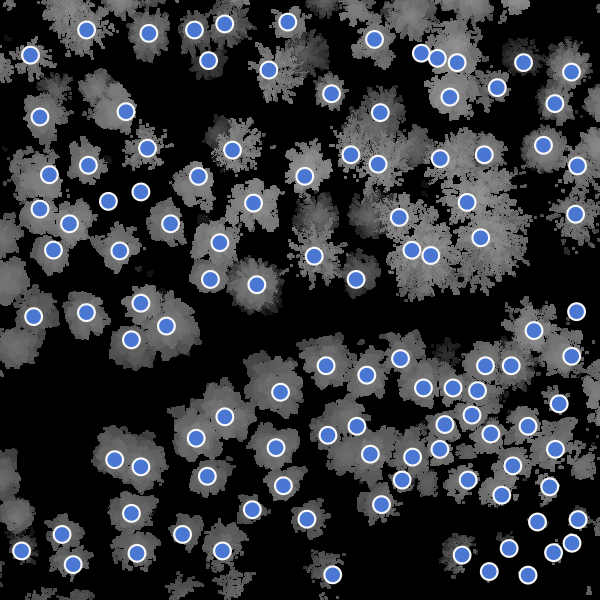}
      \caption{Peak extraction in the rasterized ALS cloud.}
  \end{subfigure}
  \hspace{0.2cm}
  \begin{subfigure}[b]{0.42\columnwidth}
      \centering
      \includegraphics[width=\textwidth, height=3.6cm]{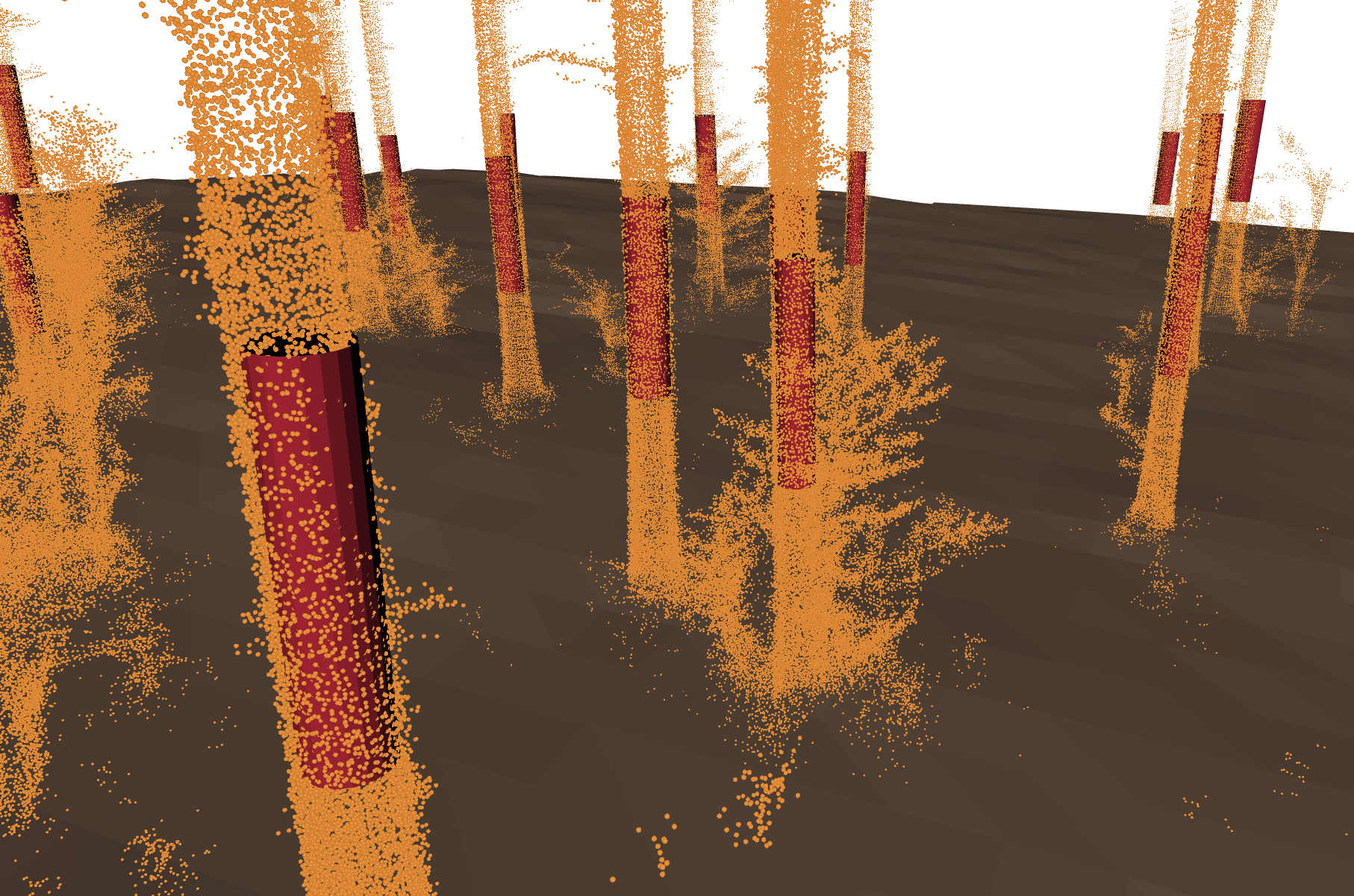}
      \caption{Tree segmentation in the MLS clouds.}
  \end{subfigure}
  \caption{Feature extraction approaches to determine the tree positions in (a) ALS and (b) MLS clouds. }
  \label{fig:cloud-features}
\end{figure}

The next step of the method is to extract features from the cropped ALS cloud
--- specifically the peaks of each tree present. We rasterize a top-down view of
the cloud by selecting the maximum height of the peak points along the vertical
direction, as also done by De Lima et al.~\cite{delima23ral}. This effectively
represents a \emph{canopy height map} (CHM) of the aerial cropped cloud (see
\figref{fig:cloud-features} (a))

We apply non-maxima suppression on the CHM to extract local peaks, which we
assume corresponds to the tree stem locations, as also considered previously by
other authors~\cite{hussein2013iros}. The output tree location corresponds to
the features of the aerial map that are used for matching, as shown in
\figref{fig:cloud-features}. 
A limitation of this approach is that for dense forest environments it is likely
to be more difficult to distinguish individual treetops due to canopy overlap.

\subsubsection{Terrestrial Feature Extraction}
\label{ssec:mls_features}
The MLS clouds are acquired by an MLS with a horizontal LiDAR and narrow
field-of-view ($\sim$\SI{30}{\degree}), inclined at a pitch angle of
\SI{20}{\degrees}. Hence, the tree canopy remains obstructed for trees around
the sensing device, making it unfeasible to generate CHM which is complete
enough to extract tree locations. To address this limitation, we use an
alternative method for terrestrial tree stem estimation, similar to Proudman et
al.~\cite{proudman2022ras} and Freißmuth et
al.~\cite{freissmuth2024realtime-trees}. 
We extract the points up to \SI{5}{\meter} above the
ground plane and apply a density-based spatial clustering algorithm
(DBSCAN~\cite{EsterKSX96}). We then fit a cylinder within a
RANSAC~\cite{fischler1981ransac} loop for each cluster, thereby obtaining the
principal axis and the center position of each tree. These tree center positions
serve as the features of the MLS cloud, as shown in \figref{fig:cloud-features}
(b).

\subsubsection{Coarse Registration using Maximum Clique}
After extracting the positions of tree trunks from both the terrestrial and
aerial clouds, we match individual trees to estimate the relative 2D
transformation between both clouds. 

To find correspondences, we use a maximum clique algorithm inspired by Bailey et
al.~\cite{baileyNRD00}. We first build an aerial graph $\mathcal{G}_{\text{a}}$
by using the tree locations from the CHM as the vertices, resulting in a
\emph{completely connected} graph. Similarly, we create a complete terrestrial
graph $\mathcal{G}_{\text{t}}$ using the estimated tree positions estimated in
\secref{ssec:mls_features}. 

Both graphs are used to construct a \emph{correspondence graph} $\mathcal{C}$
between the aerial and local terrestrial graph~\cite{chen1998unifying}.
Technically, this correspondence graph is formed by taking the Cartesian product
of the vertices from $\mathcal{G}_{\text{a}}$ and $\mathcal{G}_{\text{t}}$. For
example, given $u, u' \in \mathcal{G}_{\text{a}}$ and $v, v' \in
\mathcal{G}_{\text{t}}$, $\mathcal{C}$ has as vertices $(u, v)$, $(u,v')$,
$(u',v)$ and $(u',v')$. The edges of $\mathcal{C}$ encode pairwise consistency:
an edge between vertices $(u, v)$ and $(u', v')$ exists if and only if $|d(u,
u') - d(v, v')| < \tau $, where $d(\cdot)$ is a distance metric between vertices
of the graphs, and $\tau$ is a tolerance threshold.

The graph correspondence problem between $\mathcal{G}_{\text{a}}$ and
$\mathcal{G}_{\text{t}}$ then reduces to finding the maximum clique of the
correspondence graph $\mathcal{C}$~\cite{BarrowB76}. This allows us to obtain
matches between the 2D tree positions of the aerial and terrestrial clouds, used
to solve for the relative planar pose using Umeyama's method~\cite{umeyama}. As
previously mentioned, by cropping the input clouds, we manage to limit the
number of trees, thereby controlling the size of the correspondence graph
$\mathcal{C}$---ensuring that it remains manageable for efficient maximum clique
detection. This operation typically takes only a few seconds using the Python
library \emph{NetworkX}.

\subsubsection{Fine Registration using ICP}
The planar pose estimate from the maximum clique method, combined with the
z-offset and pitch and roll corrections from the pre-processing step provide a
coarse estimate of the relative 6-DoF transformation between the local MLS cloud
and the aerial/global frame, denoted by $\mathbf{T}_{\texttt{AM}}$.

In practice, we observed offsets of about \SI{20}{\centi\meter} in all
directions for the registered clouds at this step---particularly noticeable on
the tree trunks. Therefore, to achieve a finer registration, we conduct a final
Iterative Closest Point (ICP) refinement~\cite{BeslM92}. Finally, we filter out
those matches where the number of ICP inliers falls below a specified threshold,
as we do not require all tiles/payloads to be precisely registered to the aerial
cloud.
\subsection{Aerial-terrestrial Pose Graph Optimization}
\label{subsec:pose-graph}

\begin{figure}[t]
  \centering
  \includegraphics[width=0.95\linewidth]{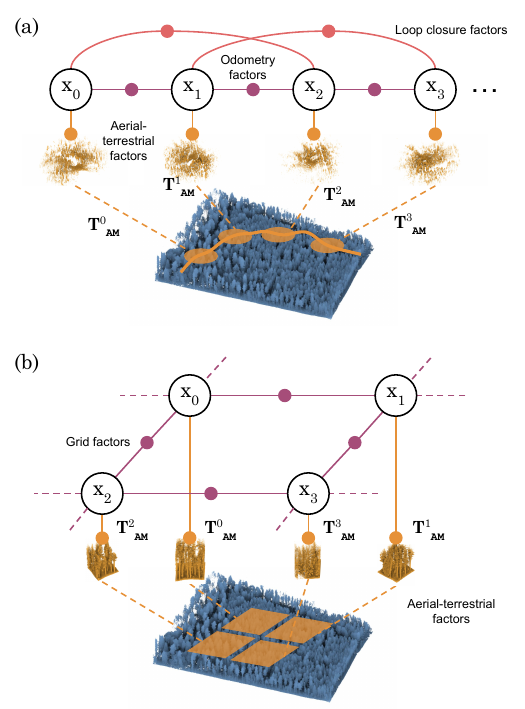}
  \caption{The factor graph formulation for aerial-terrestrial registration for our two data formats. Top: For pose graph/payload optimization we have standard odometry (purple) and loop closure factors (red) to which we add individual aerial-terrestrial prior factors (in orange). Bottom: For the tile format, the grid representation forms local constraints (purple) and the individual tiles are registered to add aerial-terrestrial prior factors (in orange).}
  \label{fig:pose-graph}
\end{figure}

To achieve global consistency between the ALS and MLS clouds, previous
approaches have proposed to solve an optimization problem that minimizes the
overall registration cost between aerial and terrestrial scans, using additional
constraints provided by cycle consistency costs~\cite{castorena2023arxiv}. Here
we explore a unified method, based on factor graph optimization, which enables
us to co-register aerial and terrestrial clouds for both the SLAM and tile
formats.

Using the approach described in~\secref{subsec:matching}, we obtain a set of
6-DoF transformations $\mathbf{T}^{l}_{\texttt{AM}}$ between the MLS clouds and
the ALS, where $l$ is the index of all the MLS clouds successfully registered
after the ICP registration step. Each transformation provides a constraint
$\mathbf{r}^{l}_{\text{AM}}$ in the factor graph, which we refer to as
\emph{Aerial-terrestrial factors}. We discard the index $l$ for simplicity.

We formulate an optimization that leverages the structure of the MLS data
format, as well as the aerial-terrestrial matches:

\begin{equation}
  \{ \mathbf{T}_{i}\} = \argmax{\underbrace{\sum{\norm{\mathbf{r_{\text{structure}}}}^{2}_{\mathbf{\Sigma}}}}_{\text{MLS structure factors}} + \underbrace{\sum{\norm{\mathbf{r_{\text{AM}}}}^{2}_{\mathbf{\Sigma}}}}_{\text{Aerial-terrestrial factors}}},
\end{equation}
where $\{ \mathbf{T}_{i} \}$ is the optimized set of poses that define the
factor graph. The technical details about the residuals are as follows.

\subsubsection{Structure factors (pose graph case)}
First, we consider the typical SLAM case in which the MLS mission is described
by a pose graph and a set of payload clouds. The structure factors are given by

\emph{odometry factors} between the poses:
\begin{equation}
  \mathbf{r_{\text{odometry}}} = \text{Log}\left(\Delta\mathbf{T}^{-1} \mathbf{T}_{i}^{-1} \mathbf{T}_{i+1} \right),
\end{equation}
where $\mathbf{T}_{i}^{-1}$,$\mathbf{T}_{i+1}$ are two consecutive poses and
$\Delta\mathbf{T}$ is the relative odometry change. Similarly, we also consider
\emph{loop closure factors} between poses $i$ and $j$ proposed by a place
recognition system:
\begin{equation}
  \mathbf{r_{\text{loop}}} = \text{Log}\left(\Delta\mathbf{T}_{i,j}^{-1} \mathbf{T}_{i}^{-1} \mathbf{T}_{j} \right).
\end{equation}
\figref{fig:pose-graph}(a) illustrates these factors.

\subsubsection{Structure factors (tiles case)}
When we use the MLS clouds in tiles format, the structure is given by binary
factors that connect a node $i$ to its neighbors $n \in \mathcal{N}$:
\begin{equation}
  \mathbf{r_{\text{grid}}} = \text{Log}\left(\Delta\mathbf{T}_{i,n}^{-1} \mathbf{T}_{i}^{-1} \mathbf{T}_{n} \right),
\end{equation}
Similarly, \figref{fig:pose-graph}(b) shows how these factors are defined
graphically.

\subsubsection{Aerial-Terrestrial factor}
Lastly, the aerial-terrestrial constraints impose the 6-DoF relative
transformations between the MLS clouds (payloads or tiles) with the ALS cloud:
\begin{equation}
  \mathbf{r_{\text{AM}}} = \text{Log}\left(\mathbf{T}_{\text{AM}}^{i} \mathbf{T}_{i}\right),
\end{equation}
This is illustrated for both cases in \figref{fig:pose-graph}.

\section{Experiments}

\subsection{Datasets}
We evaluated our approach using data collected from two field campaigns
conducted in Finland and Switzerland. The first campaign took place in Evo,
Finland, in a Southern boreal forest characterized by conifers consisting mostly
of pines and spruces. ALS point cloud data was collected using an Avartek Bower
drone covering up to \SI{730}{\meter} $\times$ \SI{540}{\meter}. MLS data was
captured by a backpack-carried Hesai XT32 LiDAR, covering areas of more than a
hectare.

The second campaign was in Stein am Rhein (SaR), Switzerland, in a mixed forest.
The aerial data was collected by a DJI M600 drone carrying a Velodyne HDL-32E
LiDAR and an RTK GNSS receiver in a single mission covering \SI{1850}{\meter}
$\times$ \SI{500}{\meter}. For the MLS we used the same setup as in Evo, and we
collected data in four different missions, covering about \SI{1}{\hectare} each.

\subsection{Deformable Registration from SLAM Missions }
\begin{figure*}[t]
  \centering
  \includegraphics[width=\linewidth]{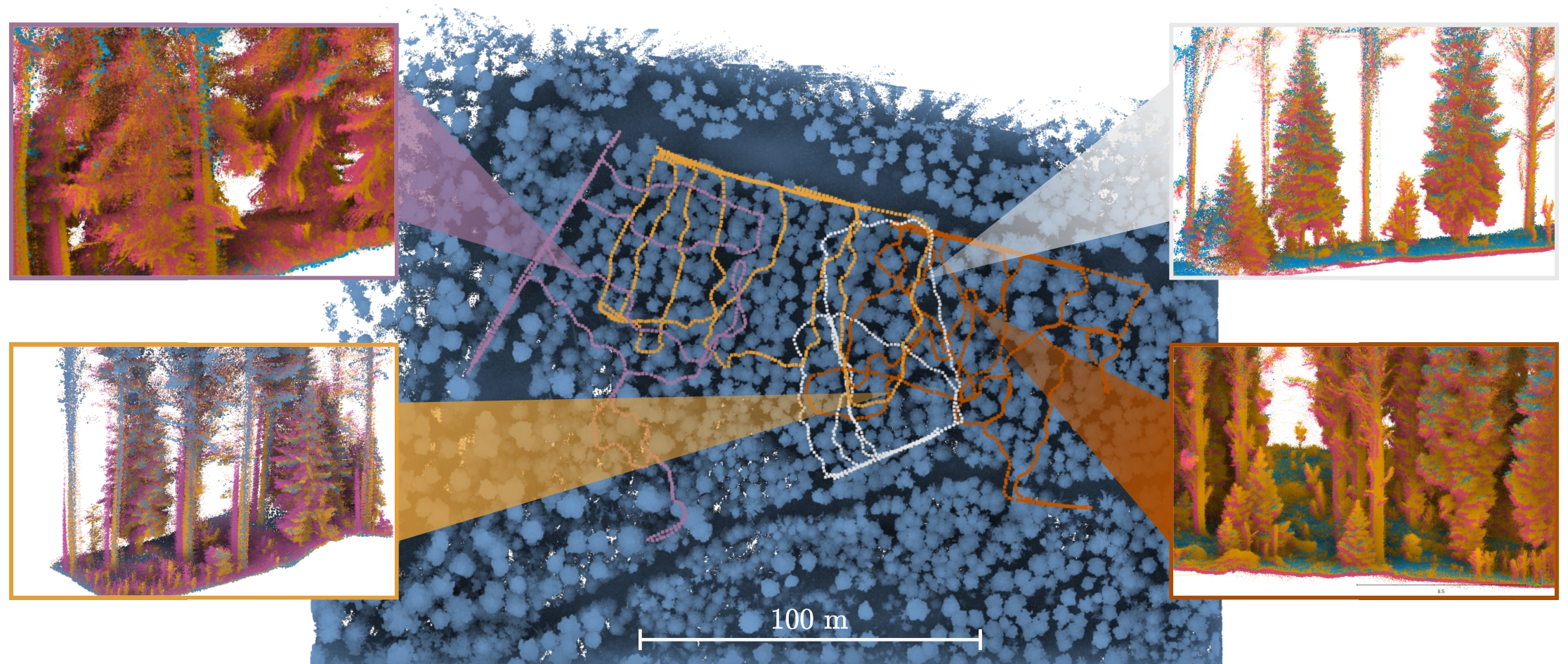}
  \caption{Illustrative examples ALS and MLS point cloud co-registrations using our proposed pipeline for four SLAM missions in Evo, Finland. MLS clouds are shown before optimization (pink), after pose graph refinement (orange), and the aerial cloud (blue). Discrepancies observed before and after the alignment highlight the corrections introduced through the addition of aerial-terrestrial constraints to the pose graph structure.}
  \label{fig:result-pose-graph}
\end{figure*}

In the first experiment, we demonstrate and evaluate our co-registration
approach across different datasets. \figref{fig:result-pose-graph} shows a
large-scale example of the co-registrations produced by our method across
multiple missions in Evo, Finland.

We register four MLS missions to an aerial map covering an overall area of about
\SI{4}{\hectare}. Each MLS mission trajectory is represented by distinct colored
lines overlaid on the ALS point cloud (in blue). The cross-sectional views at
different points on the map highlight the precise alignment accomplished through
our approach. The consistency in trunk reconstructions along the length of trees
suggests a high level of accuracy in co-registration. 

To quantitatively evaluate the advantages of the proposed pose graph refinement,
we combined all the payloads in the MLS mission before and after our
registration process. Then, we cropped the ALS cloud or the MLS cloud to obtain
the maximum overlapping subset of points, and computed the \emph{point-to-point}
error obtained before and after the pose graph refinement.

\tabref{tab:pose-graph-opt-error} presents the results obtained for all
missions. We report three different metrics: (1) point-to-point error against
the ALS cloud before and after optimization, (2) average point-wise offset after
to the optimization, and (3) average pose graph correction. 

Regarding (1), in general, we observed a decrease in the error against the ALS
cloud after the pose graph refinement --- as expected. We then inspected how
much the clouds and pose graph nodes move after optimization (2) and (3). We
report a displacement of the clouds of the order of \SI{10}{\centi\meter} on
average across all missions, while the nodes effectively moved
\SI{25}{\centi\meter} on average. The discrepancies between these numbers can be
explained by the metric used to evaluate the cloud offset (point-to-point
distance). Considering the nearest neighbors to determine the error does not
necessarily reflect how much a point moved after the optimization correction but
rather suggests a measure of the overall alignment. However, both the metrics
reflect a substantial improvement due to the optimization step, as qualitatively
shown in \figref{fig:result-pose-graph}.

\begin{table}[t]
  \centering
  \footnotesize
  \begin{tabular}{@{}lccccc@{}}
    \toprule
    \textbf{Place} & \multicolumn{1}{c}{\textbf{Mission}} &
    \multicolumn{2}{c}{\textbf{\begin{tabular}[c]{@{}c@{}}ALS-MLS error
    [m]\end{tabular}}} & \textbf{\begin{tabular}[c]{@{}c@{}}Cloud\\shift
    [m]\end{tabular}} & \textbf{\begin{tabular}[c]{@{}c@{}}Pose graph\\shift
    [m]\end{tabular}} \\
\textbf{}      & \textbf{}                            & \textbf{Pre} &
\textbf{Post}                                        & \multicolumn{1}{l}{} &
\multicolumn{1}{l}{} \\ \midrule Evo            & 1 & 0.64 & \textbf{0.61}
& 0.05 &  0.21 \\
               & 2                                    & 0.28 & \textbf{0.26} &
               0.06 &  0.23 \\
               & 3                                    & 0.39 & \textbf{0.31} &
               0.09 &  0.45 \\
               & 4                                    & \textbf{0.36} & 0.37 &
0.07 &  0.28 \\ \midrule SaR            & 1 & 0.78 & \textbf{0.39}
& 0.38 &  0.20 \\
               & 2                                    & 0.47 & \textbf{0.41} &
               0.13 &  0.38 \\ \bottomrule
    \end{tabular}
  \caption{Co-registration metrics for SLAM-based MLS missions. We evaluated the point-to-point error before and after the pose graph optimization w.r.t the ALS cloud, as well the point cloud and pose graph correction magnitude.}
  \label{tab:pose-graph-opt-error}
\end{table}

\subsection{Deformable Registration for Tile-based Data}

\begin{figure}[t]
  \centering
  \includegraphics[width=\linewidth]{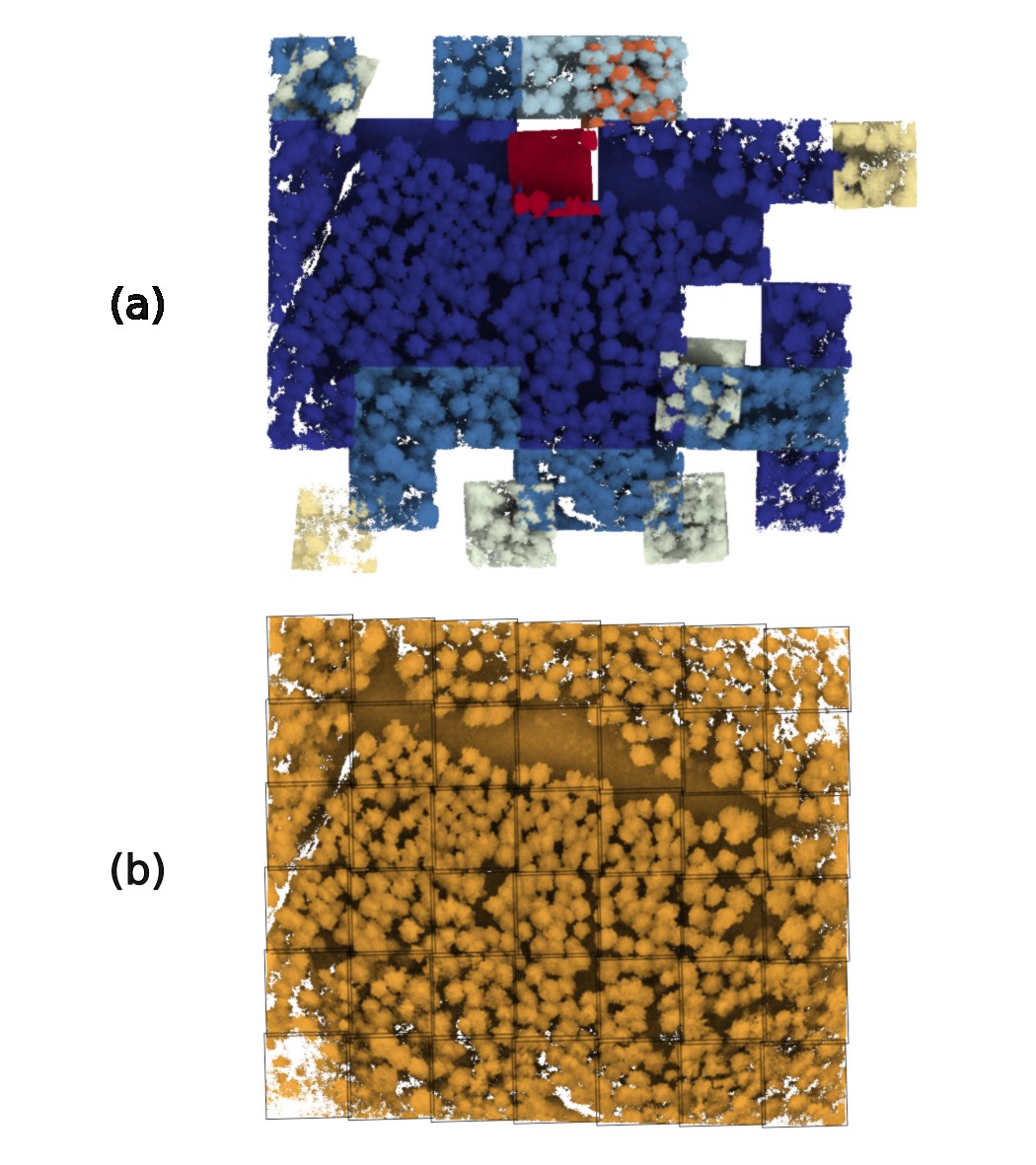}
  \caption{Deformable registration for tile-based data. Each tile measure
  \SI{20}{\meter} $\times$ \SI{20}{\meter}. (a) Registration results after the
  initial aerial-terrestrial tile-wise matching step. Blue denotes higher ICP
  matching scores, while other colors indicate lower scores. (b) Results after
  pose graph optimization, achieving alignment across all tiles despite initial
  registration failures for some of the tiles.}
  \label{fig:tiles-opt}
\end{figure}

In this experiment, we demonstrate pose graph-based optimization for tile-based
MLS data. We partitioned one of the MLS missions from the Evo dataset into
square tiles measuring \SI{20}{\meter}$\times$\SI{20}{\meter} each. We then
executed our pipeline, including the tile-wise aerial-terrestrial matching step
(see \secref{subsec:matching}), to assess the performance of registration. The
main distinction between the MLS payload data and the tiled data lies in the
density of the respective point clouds. However, since the coarse registration
step only uses sparse tree location positions, it does not impact computation
time significantly. We present the results of the initial aerial-terrestrial
registration in \figref{fig:tiles-opt}(a). Here, we colorized the tiles
individually based on the ICP fitness score obtained after the matching step,
where blue indicates a higher fitness score. Other colored tiles represent lower
fitness scores. This is often due to an insufficient number of trees present in
the corresponding tiles for unambiguous matching within the aerial cloud.

We address this limitation by leveraging the pose graph structure to propagate
corrections to tiles that initially failed to register successfully with the
aerial clouds. After optimization, guided by the prior imposed by the tile grid,
alignment is attained across all tiles, as depicted in
\figref{fig:tiles-opt}(b).

\subsection{Accuracy Analysis of Aerial-Terrestrial Transformation}
To assess the quality of the proposed aerial-terrestrial factors, we evaluated
the accuracy of the individual payload-based matches. We compared the results
obtained by our matching module against a subset of manually registered clouds.
We chose a subset of MLS payload clouds and manually registered them using
CloudCompare, selecting four corresponding points in the payload and the aerial
map, followed by ICP refinement. \tabref{tab:registration} presents the
root-mean-square error (RMSE) computed from point-to-point distances for both
the manually aligned clouds and our proposed method (\secref{subsec:matching}).
This evaluation was conducted on a subset of mission payloads for which
successful alignment was achieved using our approach. On average, we observed a
point-to-point RMSE of \SI{0.44}{\meter} for the Evo dataset and
\SI{0.31}{\meter} for the SaR dataset. The results demonstrate that our
registration method achieves accurate local alignment, with our approach showing
better point-to-point error compared to the manual alignment.

\begin{table}[t]
  \footnotesize
  \centering
  \begin{tabular}{llllll}
    \toprule
    \textbf{Area} & \textbf{Mission} & \multicolumn{2}{c}{\textbf{Manual}} &
    \multicolumn{2}{c}{\textbf{Ours}} \\ 
         &     & \textbf{mean}  & \textbf{std}  & \textbf{mean} & \textbf{std}
    \\ \midrule Evo  & 1   &   0.67      &   2.11    &   \textbf{0.41}     &
    2.20 \\ 
         & 2   &   0.31     &   0.31    &   \textbf{0.21}     & 0.30\\ 
         & 3   &   0.38     &   1.27    &   \textbf{0.35}     & 1.53\\
         & 4   &   0.40     &   0.86    &   \textbf{0.29}     & 0.94\\
         \midrule
    SaR  & 1   &   0.31     &    0.48   &   \textbf{0.24}     & 0.37 \\
         & 2   &   0.32     &    0.40   &   \textbf{0.26}     & 0.38 \\
    \bottomrule
  \end{tabular}
  \caption{Accuracy of aerial-terrestrial registration for individual payloads.
  We evaluated the point-to-point error for our registration method
  (\secref{subsec:matching}) and manual alignment as a reference.}
  \label{tab:registration}
  \end{table}

\subsection{Accessing Completeness of Co-Registered Point Clouds}
\label{subsec:completeness}
\begin{figure}[t]
  \centering
  \includegraphics[width=0.95\linewidth]{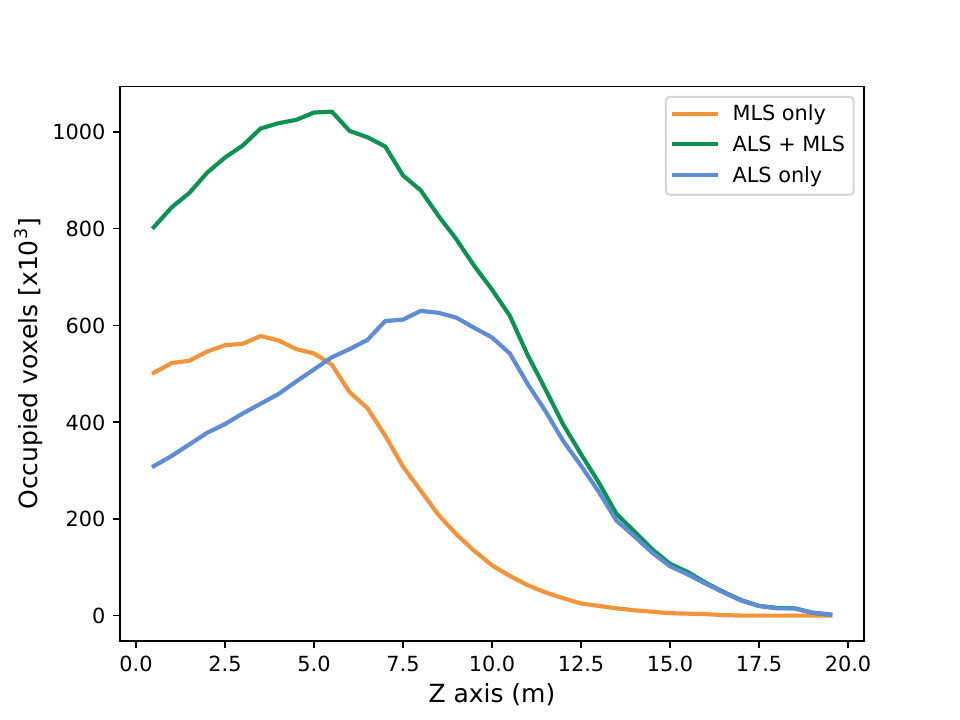}
  \caption{Completeness analysis. Voxel occupancy (with \SI{5}{\centi\meter}
  resolution) at different heights for the aerial, terrestrial, and
  co-registered point clouds. The co-registered point cloud shows a higher
  occupancy through the height of the tree.}
  \label{fig:density}
\end{figure}

Lastly, we evaluated the benefits of co-registering terrestrial and aerial
clouds in terms of \emph{completeness}. We quantified cloud completeness through
density measurements across various height intervals, utilizing a voxelized
representation to ensure occupancy evaluation rather than relying solely on raw
point density, which may vary depending on the LiDAR type. We achieved this by
voxelizing the point cloud with a resolution of \SI{5}{\centi\meter} for the
ALS, MLS, and combined clouds. \figref{fig:density} shows the results obtained
for Mission 1 of the Evo dataset. First, we observed that the distribution of
occupied voxels for ALS and MLS confirms the differences expected for both
modalities, with MLS concentrating the low-height points, and ALS the canopy and
top of the trees. The combined ALS + MLS cloud obtained with our proposed method
indeed provided more complete reconstructions, as qualitatively illustrated in
the previous experiments.

\begin{figure}[t]
  \centering
  \includegraphics[width=0.95\linewidth]{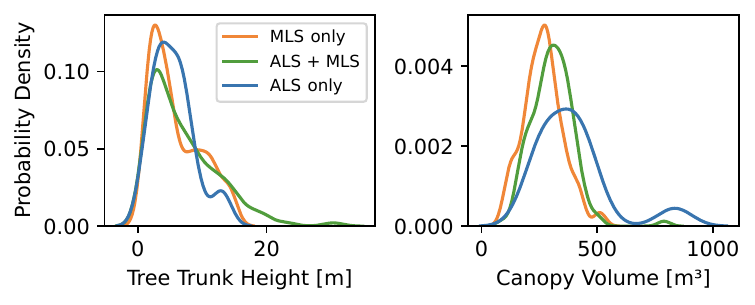}
  \caption{Distributions of tree height and canopy volume. The combined point cloud (green) shows a shift to the right, indicating greater height and canopy volume captured compared to MLS (orange) or ALS (blue) only.}
  \label{fig:pdf_heights_volumes}
\end{figure}

To further demonstrate the advantages of combining aerial and terrestrial
clouds, we ran a tree trait extraction pipeline to segment individual trees and
determine the tree height and canopy volume from ALS, MLS, and co-registered
clouds. For estimating the tree height, we used an algorithm that modeled the
stem as a set of local cylinders along the trunk---the set of estimated
cylinders was used to estimate the tree height. For the canopy volume, we used a
convex hull-based algorithm that enclosed points associated to the tree that
were not used to estimate the stem model, effectively capturing points
associated to branches and leaves. We used the volume of the hull as a proxy for
the canopy volume.

\figref{fig:pdf_heights_volumes} shows the results obtained for this experiment.
Regarding the tree height estimates, \figref{fig:pdf_heights_volumes} (a) shows
the normalized distribution of estimated heights. We observed that the combined
cloud distribution effectively covers a larger range of tree heights, observed
by the shift of the right-hand tail. This can be explained for the higher degree
of completeness of the combined clouds compared to independent ALS and MLS
clouds, which improved the robustness of the stem modeling approach for
different tree heights. We reported a similar result for the canopy volume
estimates, with the combined cloud displaying longer tails in the volume
distribution. This demonstrates that by estimating tree traits from a combined
aerial-terrestrial point cloud we can effectively produce better tree estimates
for forest inventory and monitoring.

\section{Conclusion}
\label{sec:conclusion}

This work presented a novel approach for co-registering aerial and
terrestrial point clouds. By leveraging a deformable pose graph formulation, we
achieved precise alignment by exploiting the underlying MLS SLAM representation in
a manner which the existing methods do not now. We validated our approach on diverse
forest datasets in multiple countries and demonstrate the effectiveness in
creating more accurate and complete forest inventories. This capability holds
promise for enhancing forest management and conservation efforts.

\section*{Acknowledgments}
This work was supported by the European Union's Horizon Europe under grant
agreement 101070405, a Royal Society University Research Fellowship and a
UKRI/EPSRC Programme Grant [EP/V000748/1]. The authors acknowledge the support
of PreFor Oy for providing the ALS data used in the experiments. We also thank
Janine Schweier, Holger Griess and their team at the Swiss Federal Institute for
Forest, Snow and Landscape Research (WSL) for supporting the data collection in
Stein am Rhein.

\bibliographystyle{plain_abbrv}
\bibliography{new}

\begin{thebibliography}{10}

\bibitem{baileyNRD00}
T.~Bailey, E.M. Nebot, J.~Rosenblatt, and H.F. Durrant{-}Whyte.
\newblock Data association for mobile robot navigation: {A} graph theoretic
  approach.
\newblock In {\em IEEE Int. Conf. Robot. Autom. (ICRA)}, pages 2512--2517.
  {IEEE}, 2000.

\bibitem{BarrowB76}
H.G. Barrow and R.M. Burstall.
\newblock Subgraph isomorphism, matching relational structures and maximal
  cliques.
\newblock {\em Inf. Process. Lett.}, 4(4):83--84, 1976.

\bibitem{BeslM92}
P.J. Besl and N.D. McKay.
\newblock A method for registration of 3-d shapes.
\newblock {\em {IEEE} Trans. Pattern Anal. Mach. Intell.}, 14(2):239--256,
  1992.

\bibitem{bienert2009methods}
A.~Bienert and H.G. Maas.
\newblock Methods for the automatic geometric registration of terrestrial laser
  scanner point clouds in forest stands.
\newblock {\em ISPRS Archives of the Photogrammetry, Remote Sensing and Spatial
  Information Sciences}, pages 93--98, 2009.

\bibitem{castorena2023arxiv}
J.~Castorena, L.T. Dickman, A.J. Killebrew, J.R. Gattiker, R.R. Linn, and E.L.
  Loudermilk.
\newblock Automated structural-level alignment of multi-view {TLS} and {ALS}
  point clouds in forestry.
\newblock {\em ArXiv}, abs/2302.12989, 2023.

\bibitem{chasmer2004assessing}
L.~Chasmer, C.~Hopkinson, and P.~Treitz.
\newblock Assessing the three-dimensional frequency distribution of airborne
  and ground-based lidar data for red pine and mixed deciduous forest plots.
\newblock {\em ISPRS Archives of the Photogrammetry, Remote Sensing and Spatial
  Information Sciences}, 36(8):W2, 2004.

\bibitem{chen1998unifying}
C.~Chen and D.Y. Yun.
\newblock Unifying graph-matching problem with a practical solution.
\newblock In {\em Proceedings of International Conference on Systems, Signals,
  Control, Computers}, volume~55. Citeseer, 1998.

\bibitem{dai2019automated}
W.~Dai, B.~Yang, X.~Liang, Z.~Dong, R.~Huang, Y.~Wang, and W.~Li.
\newblock Automated fusion of forest airborne and terrestrial point clouds
  through canopy density analysis.
\newblock {\em ISPRS Journal of Photogrammetry and Remote Sensing},
  156:94--107, 2019.

\bibitem{delima23ral}
L.C. de~Lima, M.~Ramezani, P.V.K. Borges, and M.~Br{\"{u}}nig.
\newblock Air-ground collaborative localisation in forests using lidar canopy
  maps.
\newblock {\em {IEEE} Robot. Autom. Lett. (RA-L)}, 8(3):1818--1825, 2023.

\bibitem{EsterKSX96}
M.~Ester, H.~Kriegel, J.~Sander, and X.~Xu.
\newblock A density-based algorithm for discovering clusters in large spatial
  databases with noise.
\newblock In E.~Simoudis, J.~Han, and U.M. Fayyad, editors, {\em Proceedings of
  the Second International Conference on Knowledge Discovery and Data Mining
  (KDD-96), Portland, Oregon, {USA}}, pages 226--231. {AAAI} Press, 1996.

\bibitem{fischler1981ransac}
M.A. Fischler and R.C. Bolles.
\newblock Random sample consensus: A paradigm for model fitting with
  applications to image analysis and automated cartography.
\newblock {\em Commun. ACM}, 24(6):381–395, 1981.

\bibitem{freissmuth2024realtime-trees}
L.~Freißmuth, M.~Mattamala, N.~Chebrolu, S.~Schaefer, S.~Leutenegger, and
  M.~Fallon.
\newblock Online tree reconstruction and forest inventory on a mobile robotic
  system.
\newblock {\em IEEE/RSJ Int. Conf. Intell. Robots Syst. (IROS)}, 2024.

\bibitem{ge2021target}
X.~Ge and Q.~Zhu.
\newblock Target-based automated matching of multiple terrestrial laser scans
  for complex forest scenes.
\newblock {\em ISPRS Journal of Photogrammetry and Remote Sensing}, 179:1--13,
  2021.

\bibitem{hilker2012simple}
T.~Hilker, N.C. Coops, D.S. Culvenor, G.~Newnham, M.A. Wulder, C.W. Bater, and
  A.~Siggins.
\newblock A simple technique for co-registration of terrestrial lidar
  observations for forestry applications.
\newblock {\em Remote sensing letters}, 3(3):239--247, 2012.

\bibitem{holopainen2013tree}
M.~Holopainen, V.~Kankare, M.~Vastaranta, X.~Liang, Y.~Lin, M.~Vaaja, X.~Yu,
  J.~Hyyppä, H.~Hyyppä, H.~Kaartinen, A.~Kukko, T.~Tanhuanpää, and P.~Alho.
\newblock Tree mapping using airborne, terrestrial and mobile laser scanning
  – a case study in a heterogeneous urban forest.
\newblock {\em Urban Forestry \& Urban Greening}, 12(4):546--553, 2013.

\bibitem{hussein2013iros}
M.~Hussein, M.~Renner, M.~Watanabe, and K.~Iagnemma.
\newblock Matching of ground-based lidar and aerial image data for mobile robot
  localization in densely forested environments.
\newblock In {\em IEEE/RSJ Int. Conf. Intell. Robots Syst. (IROS)}, pages
  1432--1437. {IEEE}, 2013.

\bibitem{lovell2003aerial}
D.S.C. J~L~Lovell, D L.B.~Jupp and N.C. Coops.
\newblock Using airborne and ground-based ranging lidar to measure canopy
  structure in australian forests.
\newblock {\em Canadian Journal of Remote Sensing}, 29(5):607--622, 2003.

\bibitem{jelavic2021autonomous}
E.~Jelavic, D.~Jud, P.~Egli, and M.~Hutter.
\newblock Towards autonomous robotic precision harvesting: Mapping,
  localization, planning and control for a legged tree harvester, 2021.

\bibitem{kankare2014accuracy}
V.~Kankare, J.~Vauhkonen, T.~Tanhuanpää, M.~Holopainen, M.~Vastaranta,
  M.~Joensuu, A.~Krooks, J.~Hyyppä, H.~Hyyppä, P.~Alho, and R.~Viitala.
\newblock Accuracy in estimation of timber assortments and stem distribution
  – a comparison of airborne and terrestrial laser scanning techniques.
\newblock {\em ISPRS Journal of Photogrammetry and Remote Sensing}, 97:89--97,
  2014.

\bibitem{haughlin2014geo}
E.N. Marius~Hauglin, Vegard~Lien and T.~Gobakken.
\newblock Geo-referencing forest field plots by co-registration of terrestrial
  and airborne laser scanning data.
\newblock {\em International Journal of Remote Sensing}, 35(9):3135--3149,
  2014.

\bibitem{murtiyoso2024}
A.~Murtiyoso, S.~Holm, H.~Riihim{\"a}ki, A.~Krucher, H.~Griess, V.C. Griess,
  and J.~Schweier.
\newblock Virtual forests: a review on emerging questions in the use and
  application of 3d data in forestry.
\newblock {\em International Journal of Forest Engineering}, 35(1):29--42,
  2024.

\bibitem{omasa2008three}
K.~Omasa, F.~Hosoi, T.M. Uenishi, Y.~Shimizu, and Y.~Akiyama.
\newblock Three-{Dimensional} {Modeling} of an {Urban} {Park} and {Trees} by
  {Combined} {Airborne} and {Portable} {On}-{Ground} {Scanning} {LIDAR}
  {Remote} {Sensing}.
\newblock {\em Environmental Modeling \& Assessment}, 13(4):473--481, November
  2008.

\bibitem{paris2017automatic}
C.~Paris, D.~Kelbe, J.~van Aardt, and L.~Bruzzone.
\newblock A novel automatic method for the fusion of als and tls lidar data for
  robust assessment of tree crown structure.
\newblock {\em IEEE Transactions on Geoscience and Remote Sensing},
  55(7):3679--3693, 2017.

\bibitem{polewski2019markerfree}
P.~Polewski, W.~Yao, L.~Cao, and S.~Gao.
\newblock Marker-free coregistration of uav and backpack lidar point clouds in
  forested areas.
\newblock {\em ISPRS Journal of Photogrammetry and Remote Sensing},
  147:307--318, 2019.

\bibitem{proudman2022ras}
A.~Proudman, M.~Ramezani, S.T. Digumarti, N.~Chebrolu, and M.~Fallon.
\newblock Towards real-time forest inventory using handheld lidar.
\newblock {\em Robotics and Autonomous Systems}, 157:104240, 2022.

\bibitem{shao2022arxiv}
J.~Shao, W.~Yao, P.~Wan, L.~Luo, J.~Lyu, and W.~Zhang.
\newblock Efficient divide-and-conquer registration of {UAV} and ground lidar
  point clouds through canopy shape context.
\newblock {\em ArXiv}, abs/2201.11296, 2022.

\bibitem{umeyama}
S.~Umeyama.
\newblock Least-squares estimation of transformation parameters between two
  point patterns.
\newblock {\em {IEEE} Trans. Pattern Anal. Mach. Intell.}, 13(4):376--380,
  1991.

\bibitem{wang2019situ}
Y.~Wang, J.~Py{\"o}r{\"a}l{\"a}, X.~Liang, M.~Lehtom{\"a}ki, A.~Kukko, X.~Yu,
  H.~Kaartinen, and J.~Hyypp{\"a}.
\newblock In situ biomass estimation at tree and plot levels: What did data
  record and what did algorithms derive from terrestrial and aerial point
  clouds in boreal forest.
\newblock {\em Remote Sensing of Environment}, 232:111309, 2019.

\bibitem{white2016remote}
J.C. White, N.C. Coops, M.A. Wulder, M.~Vastaranta, T.~Hilker, and
  P.~Tompalski.
\newblock Remote sensing technologies for enhancing forest inventories: A
  review.
\newblock {\em Canadian Journal of Remote Sensing}, 42(5):619--641, 2016.

\end{thebibliography}

\end{document}